\documentclass{article}
\usepackage{multirow}
\usepackage{arxiv}
\usepackage{stix}
\usepackage[utf8]{inputenc} % allow utf-8 input
\usepackage[T1]{fontenc}    % use 8-bit T1 fonts
\usepackage{hyperref}       % hyperlinks
\usepackage{url}            % simple URL typesetting
\usepackage{booktabs}       % professional-quality tables
\usepackage{amsfonts}  
\usepackage{amsmath} % blackboard math symbols
\usepackage{amsthm}
\usepackage{nicefrac}       % compact symbols for 1/2, etc.
\usepackage{microtype}      % microtypography
\usepackage{lipsum}		% Can be removed after putting your text content
\usepackage{graphicx}
\usepackage{subfig}
\usepackage{natbib}
\usepackage{doi}
\usepackage{tikz}
\usepackage{longtable}
\usepackage{caption}
\usepackage{booktabs}
\usepackage[utf8]{inputenc}
\usepackage{graphicx}
\usepackage{authblk}

\title{Deep Maxout Network Gaussian Process}

\author[$*$]{Libin Liang}
\author[$*$]{Ye Tian}
\author[ ]{Ge Cheng} 
\affil[ ]{Department of Statistics, Rutgers University} 

\renewcommand{\headeright}{}
\renewcommand{\undertitle}{}
\renewcommand{\shorttitle}{}
\newcommand{\cir}[1]{\tikz[baseline]{%
    \node[anchor=base, draw, circle, inner sep=0, minimum width=0.8em]{#1};}}
\newcommand{\widesim}[2][1.5]{
  \mathrel{\overset{#2}{\scalebox{#1}[1]{$\sim$}}}
}
\hypersetup{
pdftitle={DEEP Maxout Network Gaussian Process},
pdfsubject={q-bio.NC, q-bio.QM},
pdfauthor={David S.~Hippocampus, Elias D.~Striatum},
pdfkeywords={First keyword, Second keyword, More},
}

\begin{document}
\maketitle
\def\thefootnote{*}\footnotetext{These authors contributed equally to this work.}
%\def\thefootnote{*}\footnotetext{These authors 5contributed equally to this work.}

%\def\thefootnote{\arabic{footnote}}
%text text text\footnote{normal footnote}
\begin{abstract}
	%\lipsum[1]
	%asdfsdfdsfsdf
	Study of neural networks with infinite width is important for better understanding of the neural network in practical application. In this work, we derive the equivalence of the deep, infinite-width maxout network and the Gaussian process (GP) and characterize the maxout kernel with a compositional structure. Moreover, we build up the connection between our deep maxout network kernel and deep neural network kernels introduced in \citet{lee2017deep}. We also give an efficient numerical implementation of our kernel which can be adapted to any maxout rank. Numerical results show that doing Bayesian inference based on the deep maxout network kernel can lead to competitive results compared with their finite-width counterparts and deep neural network kernels introduced in \citet{lee2017deep}. This enlightens us that the maxout activation may also be incorporated into other infinite-width neural network structures such as the convolutional neural network (CNN).
\end{abstract}

% keywords can be removed

\section{Introduction}
%\lipsum[2]
%\lipsum[3]
Investigating neural networks with infinite width becomes a popular topic for two main reasons. First,  neural networks in real application typically have a super large width, which may share similar properties with infinite-width neural networks ( \citet{NEURIPS2019_0d1a9651}, \citet{NEURIPS2019_cf9dc5e4}). Second, infinite-width neural networks have analytical properties which are easier to study (\citet{DBLP:journals/corr/abs-1902-08129}). 

The infinite-width neural network as a GP was first derived in \citet{neal1996priors}. Given a proper initialization, the output of a single-layer neural network becomes a GP with the kernel depending on the nonlinearity activation. In \citet{lee2017deep}\citet{matthews2018gaussian}, the infinite-width deep neural network (DNN) as a GP was derived and the kernel has a compositional structure depending on the number of depth and the nonlinearity activation in the network. Moreover, they found that Bayesian inference with DNN kernels often outperforms their finite-width counterparts. Both \citet{neal1996priors}
and \citet{lee2017deep} studied neural networks with activations applied to single linear combinations of inputs. Following the infinite-width scheme,  \citet{novak2018bayesian} and \citet{garriga2018deep} developed the
GP kernel based on the CNN structure and \citet{DBLP:journals/corr/abs-1910-12478} developed the
GP kernel based on neural structures such as the recurrent neural network (RNN), transformer, etc. 

Although the deep maxout network is widely utilized in application, it has never been derived as a GP among previous works. The activation in maxout network  applied on multiple linear combinations of inputs (or previous layers) so that the implementation in \citet{neal1996priors} and \citet{lee2017deep} can not be directly applied here.
In this work, we formally derive the deep, infinite-width maxout network as a GP and characterize its corresponding kernel as a compositional structure. Moreover, we give an efficient method based on numerical integration and interpolation to implement the kernel. We apply the deep maxtout network kernel to Bayesian inference on MNIST and CIFAR10 datasets and it can achieve encouraging results in numerical study.

\subsection{Related Work}

In early work, \citet{neal1996priors} derived that the single-layer, infinite-width neural network is a GP. With identical, independent distributions for the parameters initialization in each neuron, the hidden units are independent and identically distributed (i.i.d.), and the output of the network will converge to a GP by Central Limite Theorem (CLT). The covariance of the output corresponding to different inputs, which is the kernel of the GP, can be identified by the covariance of a single hidden unit, which is related to the type of the nonlinearity activation and initial variance levels of biases and weights.

\citet{lee2017deep} discovered the equivalence between DNNs and GPs. They derived the equivalence by letting the width of each layer go to infinity sequentially. They also offered the derivation based on the Bayesian marginalization over intermediate layers, which did not depend on the order of limits but rely on Gaussian priors of parameters. The corresponding kernel of their deep neural network Gaussian Process (NNGP) then can be described in a compositional manner, that is, the covariance of the current layer is determined by that of the previous layer. They also proposed an efficient numerical implementation of the DNN GP kernel. 

\citet{novak2018bayesian} and \citet{garriga2018deep} studied the equivalence between CNNs with infinite channels and GPs. \citet{novak2018bayesian} introduced a Monte Carlo method to compute CNN GP kernels. \citet{garriga2018deep} derived the same equivalence by letting the channels in each layer go to infinity sequentially as \citet{lee2017deep} and calculated the kernel exactly for the infinite-channel CNN with the ReLU activation.

The neural network structures studied in \citet{neal1996priors}, \citet{lee2017deep} and \citet{matthews2018gaussian} contain activations applied to single linear combinations of inputs (or outputs from the previous layer). In contrast, the activation in deep maxout networks applied to multiple linear combinations of inputs. Thus, directly applying previous derivations and the  implementation of DNN kernels to the deep maxout network kernel is impossible. 
%\textbf{can not directly apply in Deep Maxout Network}.

%and it also brings greater challenge to the implementation of the deep maxout networks GP kernel.

%\citet{novak2018bayesian} and %\citet{garriga2018deep} studied the %equivalence between CNN with infinite %channels and GPs. %\citet{novak2018bayesian} introduces a %Monte Carlo method to compute CNN GP %kernels. \citet{garriga2018deep} derives %the equivalence by bayesian %marginalization as \citet{lee2017deep} and %calculates the kernel exactly for infinite %channels CNN with ReLU activation. 

\subsection{Summary of Contributions}
Our novel contributions in this work are as follows:

1. We derive the equivalence of deep, infinite-width maxout networks and GPs and detailedly characterize the corresponding GP kernel with a compositional structure. We further show that the deep maxout network kernel with maxout rank $q= 2$ can be transformed to the DNN kernel with the ReLU activation with proper scale modification of the input and the variance level of initialization, which builds up the connection between the deep maxout network kernel and the deep neural network kernel.

%based on bayesian marginalization over intermediate layer described in \citet{lee2017deep} 

2. We introduce an efficient
implementation of the deep maxout network GP kernel based on numerical integration and interpolation method, which is with high precision and can be adapted to any maxout rank.

3. We conduct numerical experiments on MNIST and CIFAR10 datasets, showing that Bayesian inference with the deep maxtout network kernel often outperforms their finite-width counterparts and it can also achieve competitive results compared with DNN kernels in \citet{lee2017deep}, especially, in the more challenging dataset CIFAR10, which shows great potential to incorporate the maxout activation into other infinite-width neural network structures such as CNNs. 

\section{Review of the Maxout Network}
\label{sec:headings}
The maxout network was initially introduced in \citet{pmlr-v28-goodfellow13}. %to explore benefits of dropout. 
A maxout activation can be formalized as follows: given an input $\bold{x} \in \mathbb{R}^{d}$ ($\bold{x}$ may be the input vector or a hidden layer's state), suppose the number of linear sub-units combined by a maxout activation (maxout rank) is $q$ (in general, $ q \ll d$), a maxout activation first computes $q$ linear feature mappings $\mathbf{z} \in \mathbb{R}^{q}$, where
\begin{equation}
    \begin{aligned}
        z_{i} = \bold{w}^{\top}_{i} \bold{x} + b_{i}, \quad \bold{w}_{i} \in \mathbb{R}^{d}, \quad i \in [q].
    \end{aligned} \label{1}
\end{equation}
Afterwards, the output of the maxout hidden unit, $h_{maxout}$, is given as the maximum over the $q$ feature mappings:
\begin{equation}
    \begin{aligned}
        h_{maxout}(\bold{x}) = \text{max}\{ z_{i} \}_{i = 1}^{q}.
    \end{aligned}\label{2}
\end{equation}
Thus, suppose $\bold{w}_{i}'s$ are linearly independent, the maxout activation can be interpreted as performing a pooling operation over a $q$-dimensional affine space: 
\begin{equation}
    \begin{aligned}
        \mathcal{A} = \bold{b} + \mathcal{W},
    \end{aligned}
\end{equation}
where $\bold{b}$ is the bias vector and $\mathcal{W}=span(\{ \bold{w}_{i}\}_{i = 1}^{q})$, which is an affine transformation of the input space $\mathbb{R}^{d}$. The cross-channel max pooling activation selects the maximal output to be fed into the next layer. Such a special structure makes the maxout activation quite different from conventional activation units, which only work on an one-dimensional affine space. 

For a fully-connected deep maxout network with L hidden layers, suppose the $l_{th}$ layer has $N_{l}$ hidden units with the maxout rank q, following previous notations, the output of the the $l_{th}$ layer $out^{l}$ would be 
\begin{equation}
 out^{l} = \{ h^{l,j}_{maxout}(out^{l-1}) \}_{j = 1}^{N_{l - 1}},  
\end{equation}
where the superscript $l, j$ of $h^{l, j}_{maxout}$ represents the $j_{th}$ maxout unit in the $l_{th}$ layer, and $h^{l, j}_{maxout}$ has the structure defined in \eqref{1} and \eqref{2} with \eqref{1} adapted to $out^{l-1}$:
\begin{equation}
    \begin{aligned}
        z^{l, j}_{i} = (\bold{w}^{l, j}_{i})^{\top} \bold{x} + b^{l, j}_{i}, \quad \bold{w}^{l, j}_{i} \in \mathbb{R}^{N_{l - 1}}, \quad i \in [q].
    \end{aligned}
\end{equation}

\section{Deep Maxout Network with Infinite Width}

%We first derive the correspondence between GPs and infinitely wide neural networks in the single-layer and then extend to the multi-layer case.

\subsection{Notation}
\label{subsection:3.1}
%\label{sec:others}
%\subsection{Citations}
%Citations use \verb+natbib+. The documentation may be found at
%\begin{center}
%	\url{http://mirrors.ctan.org/macros/latex/contrib/natbib/natnotes.pdf%}
%\end{center}

%Consider a fully connected maxout network with $L$ hidden layers, which are of width $N_{l}$ (for layer $l$). Let $d_{\text{in}}$ denote the input dimension, $d_{\text{out}}$ the output dimension, $\bold{x} \in \mathbb{R}^{d_{in}}$ the input vector and $\bold{z} \in \mathbb{R}^{d_{out}}$ the output of the network. For the $i$th unit in the $l$th layer, the post-affine and pose-max transformation of the $j$th sub-component are denoted $z_{i,j}^{l}$ and $x_{i}^{l}$, respectively. We will refer to these as the pre- and
%post-max. For the input layer, we let $x_{i}^{0} \equiv \bold{x}_{i}$, the $i$th input and $\bold{x}^{\gamma} = [x^{\gamma}_{1} \cdots, x^{\gamma}_{d_{in}}]^{\top}$, $z_{i,j}^{1}(\bold{x}^{\gamma})$, a specific output of input $\gamma$. Let $b_{i,j}^{l}$ denote the bias parameter of the $j$th sub-component of $i$th unit of the $l$th layer, and $W_{i,j,k}^{l}$ the weight parameter of the corresponding $k$th input. All weight parameters are independent and randomly drawn from distributions with mean 0 and variances $\sigma_{w}^{2}/N_{l}$; and all bias parameters $ \underset{\sim}{i.i.d} N(0, \sigma_{b}^{2})$. $\mathcal{GP}(\mu, K)$ denotes a Gaussian process
%with mean and covariance functions $\mu(\cdot)$, $K(\cdot, \cdot)$, respectively.

Consider a fully connected maxout network with $L$ hidden layers, which are of width $N_{l}$ (for the $i_{th}$ layer). Let $d_{\text{in}}$ denote the input dimension, $d_{\text{out}}$ the output dimension, $\bold{x} \in \mathbb{R}^{d_{in}}$ the input vector and $\bold{z} \in \mathbb{R}^{d_{out}}$ the output of the network. For the $i_{th}$ unit in the $l_{th}$ hidden layer, let $x_i^l$ denote the return after the maxout activation and let $z_{i,j}^{l-1}$ denote the $j_{th}$ linear affine combination before the max-out activation. For the input layer, we let $x_{i}^{0} \equiv \bold{x}_{i}$ (the $i_{th}$ position of $\bold{x}$) so that we have $N_0=d_{in}$. And let $q$ denote the maxout rank of the activation. For the $i_{th}$ unit in the $l_{th}$ hidden layer, let $b_{i,j}^{l-1}$ denote the bias parameter of the $j_{th}$ sub-component of the $i_{th}$ unit, and $W_{i,j,k}^{l-1}$ the weight parameter of the corresponding $k_{th}$ position of the input. For
the output layer, let $b_i^L$ denote the bias parameter of the $i_{th}$ output and $W_{i,k}^L$ denote the weight parameter
of the corresponding $k_{th}$ position of the input. Suppose weight parameters are independently sampled from $N(0, \sigma_{w}^{2}/N_{l})$; bias parameters are independently sampled from $N(0, \sigma_{b}^{2})$, which are also independent from weight parameters. 
%In fact, we can choose different $\sigma_{b}^{l}$, $\sigma_{w}^{l}$ for layer $l$, which we will also use when it is needed to. 
Let $\mathcal{GP}(\mu, K)$ denote a Gaussian process
with mean and covariance functions $\mu(\cdot)$, $K(\cdot, \cdot)$, respectively. And we define function $F_q(\cdot)$ as
\begin{equation}
\begin{aligned}
 F_q(\rho)&=E[max\{h_1,...,h_q\}*max\{h_1^\prime,...,h_q^\prime\}],~~~\text{for }\forall \rho \in [-1,1],\\
 \label{4}
\end{aligned}
\end{equation}
where $\{h_i\}_{i=1,..,q} \widesim{i.i.d} N(0,1)$, $\{h_i^\prime\}_{i=1,..,q} \widesim{i.i.d} N(0,1)$ and $Cor(h_l, h_m^\prime)=\bold{1}_{l=m}*\rho$. 

Here $\bold{1}_{l=m}=
%\begin{equation*}
%\label{eq6}
\left\{
\begin{aligned}
1 & , & \text{if}~l=m \\
0& , & l\neq m
\end{aligned}
\right.$
%\end{equation*}$

\subsection{Deep, Infinite Width Maxout Networks as Gaussian Processes}

We first derive the equvailence between GPs and infinite-width maxout networks in the single-layer case and then extend the equivalence to the multi-layer case.

\subsubsection{Single-layer Maxout Networks as Gaussian Processes}
The $i_{th}$ component of the network output of the input $\bold{x}$ is computed as 
\begin{equation}
    \begin{aligned}
    z_i(\bold{x}) & = b_{i}^{1} + \sum_{k = 1}^{N_{1}} W^{1}_{i,k} x_{k}^{1} (\bold{x}) ~~~~i=1,...,d_{out},\\
    x_{k}^{1}(\bold{x}) & = \max_{m \in [q]} (z^0_{k,m}(\bold{x})),\\
    z^0_{k,m}(\bold{x})&=b_{k,m}^{0} + \sum_{r = 1}^{d_{in}} W^0_{k,m,r}x_{r}.\\
    \end{aligned} \label{5}
\end{equation}
Since weight parameters and bias parameters are all i.i.d. respectively, for the fixed input $\bold{x}$, $\{x_{k}^{1}(\bold{x})\}_{k=1,...,N_1}$ are i.i.d random variables. Let $\omega_{i,j,k}$ denote $W^{1}_{i,j,k} x_{k}^{1} (\bold{x})$, then, $\omega_{i,j,k}$, $ k \in [N_{1}]$ are i.i.d, too. And we have
\begin{equation}
    \begin{aligned}
    \mathbf{E}[\omega_{i,j,k}] & =  0, \\
    Var(\omega_{i,j,k})
    & = \frac{\sigma_{w}^{2} \cdot \mathbf{E}[(x_{k}^{1} (\bold{x}))^{2})]}{ N_{1}}.  \\
    \end{aligned}
\end{equation}
By CLT,
\begin{equation}
    \begin{aligned}
    \sum_{k = 1}^{N_{1}} W^{1}_{i,j,k} x_{k}^{1} (\bold{x})\xrightarrow[N_{1} \rightarrow +\infty]{\mathcal{D}} N(0, \sigma_{w}^{2} \cdot \mathbf{E}[(x_{k}^{1} (\bold{x}))^{2}]). 
    \end{aligned}   
\end{equation}

By Continuous Mapping Theorem (CMT), \begin{equation}
    \begin{aligned}
    z_i (\bold{x})\xrightarrow[N_{1} \rightarrow +\infty]{\mathcal{D}} N(0, \sigma_{b}^{2} + \sigma_{w}^{2} \cdot \mathbf{E}[(x_{k}^{1} (\bold{x}))^{2})]). 
    \end{aligned}   
\end{equation}

Moreover, given a finite input set $\{ \bold{x}^{\gamma_{1}}, \cdots, \bold{x}^{\gamma_{n}} \}$, Let $\omega_{i,j,k}(r)$ denote $W^{1}_{i,j,k} x_{k}^{1} (\bold{x}^{\gamma_{r}})$, n-dimensional random vectors $\{\bold{v_{k}} = (\omega_{i,j,k}(1), \cdots, \omega_{i,j,k}(n))^{\top}\}_{k=1,...,N_1}$ are independent identically-distributed random vectors with mean 0, and covariance matrix $\Sigma(n) / N_{1}$, the $(i,j)_{th}$ element of $\Sigma(n)$, $\Sigma_{i,j}(n) = \sigma_{w}^{2} \cdot \mathbf{E}[x_{k}^{1}(\bold{x}^{\gamma_{i}}) \cdot x_{k}^{1}(\bold{x}^{\gamma_{j}})]$.  Therefore, combining the multi-dimensional CLT and CMT as above, we get 
\begin{equation}
\begin{bmatrix}
z_i (\bold{x}^{\gamma_{1}}) \\
z_i (\bold{x}^{\gamma_{2}}) \\
\vdots \\
z_i (\bold{x}^{\gamma_{n}})
\end{bmatrix}
\xrightarrow[N_{l} \rightarrow +\infty]{\mathcal{D}} N(0, \sigma_{b}^{2} *1_{n\times n} + \Sigma(n)),
\end{equation}
where $1_{n\times n}$ is the $n \times n$ matrix whose elements are all 1, which is exactly the definition of a GP. We conclude that $z_i \sim \mathcal{GP}(0, K^{1})$, where
\begin{equation}
    K^{1}(\bold{x}, \bold{x}^{'}) = \mathbf{E}[z_i(\bold{x}) z_i(\bold{x}^{'})] = \sigma_{b}^{2} + \sigma_{w}^{2}\mathbf{E}[x_{k}^{1}(\bold{x}) x_{k}^{1}(\bold{x}^{'})] = \sigma_{b}^{2} + \sigma_{w}^{2} C(\bold{x}, \bold{x}^{'}), \label{10}
\end{equation}
where $C(\bold{x}, \bold{x}^{'}) \equiv \mathbf{E}(x_{k}^{1}(\bold{x}) \cdot x_{k}^{1}(\bold{x}^{\prime}))$. Note that $C(\bold{x}, \bold{x}^{'})$ is independent of $k$. Moreover, we have
\begin{equation}
    \begin{aligned}
    &\{z^0_{km}(\bold{x})\}~\widesim{i.i.d}~N(0, \sigma_b^2+\sigma_w^2\frac{||\bold{x}||^2}{d_{in}}),\\
    &\{z^0_{km}(\bold{x}^\prime)\}~\widesim{i.i.d}~N(0, \sigma_b^2+\sigma_w^2\frac{||\bold{x}^\prime||^2}{d_{in}}),\\
    &Cov(z_{kl}^0(\bold{x}),~ z_{km}^0(\bold{x}^\prime))=\bold{1}_{l=m}*(\sigma_b^2+\sigma_w^2\frac{<\bold{x},\bold{x}^\prime>}{d_{in}}),
    \end{aligned} \label{11}
\end{equation}
%If we normalized all the input
%such that %$\frac{||x||^2}{d_{in}}=\frac{|%|x^\prime||^2}{d_{in}}=1$, then %we have
where $<\bold{x},\bold{x}^\prime>$ is the Euclidean inner product of $\bold{x}$ and $\bold{x}^\prime$.

Then we have
\begin{equation}
    \begin{aligned}
    C(\bold{x}, \bold{x}^\prime)&= E[x_k^1(\bold{x}), x_k^1(\bold{x}^\prime)]\\ &= E[max\{z_{k1}^0(\bold{x}),...,z_{kq}^0(\bold{x})\}*max\{z_{k1}^0(\bold{x}^\prime),...,z_{kq}^0(\bold{x}^\prime)\}] \\
    &= \sqrt{\sigma_b^2+\sigma_w^2\frac{||\bold{x}||^2}{d_{in}}}*\sqrt{\sigma_b^2+\sigma_w^2\frac{||\bold{x}^\prime||^2}{d_{in}}}*F_q\bigg(\frac{\sigma_b^2+\sigma_w^2\frac{<\bold{x},\bold{x}^\prime>}{d_{in}}}{\sqrt{\sigma_b^2+\sigma_w^2\frac{||\bold{x}||^2}{d_{in}}}\sqrt{\sigma_b^2+\sigma_w^2\frac{||\bold{x}^\prime||^2}{d_{in}}} }\bigg).
    \end{aligned}
\end{equation}

Note that, if we define
%\begin{equation}
%    K^0(x,x^\prime) = \sigma_b^2+\sigma_w^2\frac{<x,x^\prime>}{d_{in}} \label{13}
%\end{equation}
\begin{equation}
    K^0(\bold{x},\bold{x}^\prime) = \frac{<\bold{x},\bold{x}^\prime>}{d_{in}}, \label{13}
\end{equation}
then we have
\small
\begin{equation}
    K^1(\bold{x},\bold{x}^\prime) = \sigma_b^2+\sigma_w^2*\sqrt{\sigma_b^2+\sigma_w^2K^0(\bold{x},\bold{x})}*\sqrt{\sigma_b^2+\sigma_w^2K^0(\bold{x}^\prime,\bold{x}^\prime)}*F_q(\frac{\sigma_b^2+\sigma_w^2K^0(\bold{x},\bold{x}^\prime)}{\sqrt{\sigma_b^2+\sigma_w^2K^0(\bold{x},\bold{x})}*\sqrt{\sigma_b^2+\sigma_w^2K^0(\bold{x}^\prime, \bold{x}^\prime)}}). \label{14}
\end{equation}
\normalsize

\subsubsection{Multiple-layer Maxout Networks as Gaussian Processes}

For a fully connected maxout network with $L$ hidden layers, the structure can be described as the following with a compositional manner. 

For the output layer, the $i_{th}$ component, $z_i(x)$, can be computed by
\begin{equation}
    \begin{aligned}
        z_i(\bold{x})&=b_i^L + \sum_{k=1}^{N_L} W_{i,k}^Lx_{k}^L(\bold{x}) \\
      x_k^L(\bold{x})&=max(z_{k,m}^{L-1}(\bold{x}))
    \end{aligned} \label{15}
\end{equation}

For the $l_{th}$ hidden layer, the $j_{th}$ linear affine combination in the $i_{th}$ unit, $z_{i,j}^l(x)$, can be computed by
\begin{equation}
    \begin{aligned}
        z_{i,j}^l(\bold{x}) & = b_{i,j}^{l} + \sum_{k = 1}^{N_{1}} W^{l}_{i,j,k} x_{k}^{l} (\bold{x}), \\
    x_{k}^{l}(\bold{x}) & = \max_{m \in [q]} (z^{l-1}_{k,m}(\bold{x})),
    \end{aligned} \label{16}
\end{equation}
for $l=0$,
\begin{equation}
    \begin{aligned}
       z^0_{k,m}(\bold{x})&=b_{k,m}^{0} + \sum_{r = 1}^{d_{in}} W^0_{k,m,r}x_{r}. 
    \end{aligned} \label{17}
\end{equation}

In analogy to the compositional structure in \eqref{14}, for $\forall l\in \mathbb{N}_+$, we can define a compositional kernel as below:
\small
\begin{equation}
    \begin{aligned}
        K^l(\bold{x},\bold{x}^\prime) = \sigma_b^2+\sigma_w^2*\sqrt{\sigma_b^2+\sigma_w^2K^{l-1}(\bold{x},\bold{x})}*\sqrt{\sigma_b^2+\sigma_w^2K^{l-1}(\bold{x}^\prime,\bold{x}^\prime)}*F_q(\frac{\sigma_b^2+\sigma_w^2K^{l-1}(\bold{x},\bold{x}^\prime)}{\sqrt{\sigma_b^2+\sigma_w^2K^{l-1}(\bold{x},\bold{x})}*\sqrt{\sigma_b^2+\sigma_w^2K^{l-1}(\bold{x}^\prime, \bold{x}^\prime)}}), \label{18}
    \end{aligned}
\end{equation}
\normalsize

where $K^0(\bold{x},\bold{x}^\prime)$ is defined in \eqref{13}.

The following theorem justifies the equavalence between infinite-width Multiple-layer Maxout Networks and GPs.
\newtheorem{theorem}{Theorem}
\begin{theorem}
For a fully connected maxout network with $L$ hidden layers with the structure defined in \eqref{15}, \eqref{16}, \eqref{17}, weights and biases intialized as described in \ref{subsection:3.1}, for any finite input set $\{\bold{x}^{\gamma_j}\}_{j=1}^B$, the distribution of the output $\{z_i(\bold{x}^{\gamma_j})\}_{j=1}^B$ will be a $\mathcal{GP}(0, K^L)$, if $N_1\to\infty,...,N_L\to \infty$, where the kernel $K^L$ is defined as \eqref{18}.
\end{theorem}

$\newline$
The proof of \textbf{Theorem 1} is in \ref{proof}.

\subsubsection{Bayesian Inference of Infinite-width, Deep Maxout Networks Using Gaussian Process Priors}

With the deep maxout network gaussian process (MNNGP) defined above, we can apply the Bayesian inference to make predictions for testing data. Suppose that we have $B_1$ training data $X_{train}=\{\bold{x}^{\gamma_j}\}_{j=1}^{B_1}$, $Y_{train}=\{y^{\gamma_j}\}_{j=1}^{B_1}$ and $B_2$ testing data $X_{test}=\{\bold{x}^{\eta_j}\}_{j=1}^{B_2}$, the true outcome are $t_{train}=\{t^{\gamma_j}\}_{j=1}^{B_1}$ and $t_{test}=\{t^{\eta_j}\}_{j=1}^{B_2}$. In the Bayesian training, we assume that 
\begin{equation}
    \begin{aligned}
        y^{\gamma_j} &= t^{\gamma_j} + \epsilon^{\gamma_j}, \\
        \begin{pmatrix}t_{train}\\t_{test}\end{pmatrix} &\sim N\bigg(0_{B_1+B_2}, \begin{pmatrix} K_{X_{train},X_{train}}&K_{X_{test},X_{train}}\\K_{X_{train},X_{test}}&K_{X_{test},X_{test}} \end{pmatrix}\bigg),
    \end{aligned}
\end{equation}
where $\epsilon^{\gamma_j} \widesim{i.i.d} N(0,\sigma_\epsilon^2)$,
$K_{X_{train},X_{train}}\in \mathbb{R}^{B_1\times B_1}$, $K_{X_{test},X_{train}}=K_{X_{train},X_{test}}^T\in \mathbb{R}^{B_1\times B_2}$ and $K_{X_{test},X_{test}}\in \mathbb{R}^{B_2\times B_2}$. Following the derivation in \citet{lee2017deep}, we have $t_{test}|X_{train}, Y_{train}\sim N(\mu, \Sigma)$, where
\begin{equation}
    \begin{aligned}
      \mu&=K_{X_{test},X_{train}}(K_{X_{train},X_{train}}+\sigma_\epsilon^2 I_n)^{-1}t_{train}, \\
      \Sigma&= K_{X_{test},X_{test}} - K_{X_{test},X_{train}}(K_{X_{train},X_{train}}+\sigma_\epsilon^2)^{-1}K_{X_{train},X_{test}}.\label{20}
    \end{aligned}
\end{equation}

%\begin{equation}
%    \begin{aligned}
%       (K_{11})_{ij}&=K^L(\bold{x}^{\gamma_i}, \bold{x}^{\gamma_j})~~~~~i,j=1,...,B_1 \\
%       (K_{12})_{ij}&=K^L(\bold{x}^{\gamma_i}, \bold{x}^{\eta_j}) ~~~~~i=1,...,B_1,~j=1,...,B_2  \\
%       (K_{22})_{ij}&=K^L(\bold{x}^{\eta_i}, \bold{x}^{\eta_j})~~~~~i,j=1,...,B_2
%    \end{aligned}
%\end{equation}

%(mention use cholesky decomposition and schur decomposition, term MNNGP)

\subsection{Relation to DNN Kernel with ReLU Activation}

DNNs with the ReLU activation is a speical case of deep maxout networks. We can also easily rewrite a maxout network with the maxout rank q = 2 into a DNN with the ReLU activation. So, for q = 2, in the finite-width regime, deep maxout networks and DNNs with the ReLU activation can be transformed to each other easily. Interestingly, we can derive that  our deep maxout network kernel with the maxout rank equals to 2 (also with proper initialization and proper input scaling) is equivalent to the DNN Kernel with the ReLU activation defined in \cite{lee2017deep}, which is summariezd in the following proposition.
\newtheorem{prop}{Proposition}
\begin{prop}
\label{prop}
Let $K^{l}_{RL}(\bold{x}, \bold{x^{\prime}};\Tilde{\sigma}_{b}, \Tilde{\sigma}_{w})$ denote covariance functions of the infinite-width, $L-$layer Deep neural network with the ReLu activation defined in \cite[Section 2]{lee2017deep} with bias parameter $\Tilde{\sigma}_{b}$ and weight parameter $\Tilde{\sigma}_{w}$, let $\sigma_{b} = \Tilde{\sigma}_{b}$, $\sigma_{w} = \frac{\Tilde{\sigma}_{w}}{\sqrt{2}}$,  then for covariance functions $K^{l}(\bold{x}, \bold{x^{\prime}})$ defined in \eqref{18}, we have
\begin{equation}
   K^{l}(\bold{x}, \bold{x^{\prime}}) = K^{l}_{RL}(\frac{\bold{x}}{\sqrt{2}}, \frac{\bold{x}^\prime}{\sqrt{2}};\Tilde{\sigma}_{b}, \Tilde{\sigma}_{w}), \text{ for } l \in \mathbb{N}_{+}. \label{19} 
\end{equation}
\end{prop}

The proof of \textbf{Proposition 1} is in \ref{pp}. 

From \textbf{Proposition 1}, we know that with proper initialization and proper scale of the input, the initial distribution of the output of the deep maxout network with  maxout rank equals to 2 and the output of the deep neural network with the ReLU activation are the same if the width of hidden layers goes to infinity. That is, in the infinite-width regime, the initial output of the deep maxout network with $q=2$ and the deep neural network with the ReLU activation can also be transformed to each other, which implies the tight connection between deep maxout networks with $q=2$ and deep neural networks with the ReLU activation.

\section{Implementation of Deep Maxout Network Kernel}

To implement the deep maxout network kernel, it suffices to implement the function $F_q(\rho),~\rho \in [-1,1]$ efficiently. The idea of the implementation is that we first evaluate the values of $F_q(\cdot)$ for certain grids $\{\rho_i\}_{i=1,...,n_\rho}\subset [-1,1]$. Then, given any other $\rho^*\in[-1,1]$, $F_q(\rho^*)$ will be evaluated by interpolation method based on the outputs from the grids. Our numerical implementation is easy to be adapted to any maxout rank $q$.

The details of evaluation of $F_q(\rho)$ for a specific $\rho\in[-1,1]$ are in \ref{implementation}.

\subsection{Numericial Implementation Example}
\begin{figure}[htp]
    \centering
    \includegraphics[width=7cm]{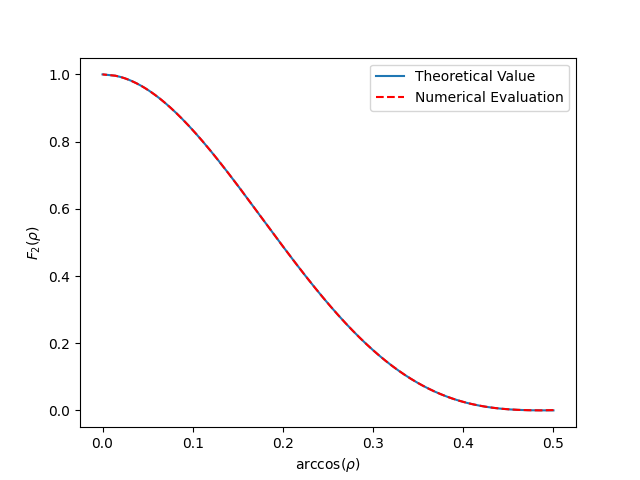}
    \caption{Theoretical Values VS Numerical Evaluations of $F_2(\rho)$. 
    The plot shows a good match between the numerical implementation and theoretical values, which demonstrates the high precision of our numerical implementation.}% We can see the numerical evaluation matches the theoretical value well.}
    \label{fig1}
\end{figure}

Based on \eqref{34} in \ref{pp}, for the maxout rank $q=2$, we have, for $\forall \rho \in [-1,1]$
\begin{equation}
    F_2(\rho)=2*E[max\{0,k_1\}*max\{0,k_2\}],
\end{equation}
where $\begin{pmatrix}k_1\\k_2\end{pmatrix}\sim N_2\bigg(0,\begin{pmatrix}1&\rho\\\rho&1\end{pmatrix}\bigg)$. From \citet{NIPS2009_5751ec3e}, we have
\begin{equation}
    E[max\{0,k_1\}*max\{0,k_2\}]=\frac{1}{2\pi}*(\sin(\arccos(\rho))+(\pi-\arccos(\rho))*\rho).
\end{equation}
We compare numerical evaluations and theoretical values for $F_2(\rho)$ in Figure \ref{fig1}.
\vspace{-1em}

%From the plot above, the numerical evaluation of $F_2(\rho)$ matches the theoretical value well.

\section{Numerical Experiment}
%See awesome Table~\ref{tab:table}.

%The documentation for \verb+booktabs+ (`Publication quality tables in %LaTeX') is available from:
%\begin{center}
%	\url{https://www.ctan.org/pkg/booktabs}
%\end{center}

%\begin{table}
%	\caption{Sample table title}
%	\centering
%	\begin{tabular}{lll}
%		\toprule
%		\multicolumn{2}{c}{Part}                   \\
%		\cmidrule(r){1-2}
%		Name     & Description     & Size ($\mu$m) \\
%		\midrule
%		Dendrite & Input terminal  & $\sim$100     \\
%		Axon     & Output terminal & $\sim$10      \\
%		Soma     & Cell body       & up to $10^6$  \\
%		\bottomrule
%	\end{tabular}
%	\label{tab:table}
%\end{table}

%\subsection{Lists}
%\begin{itemize}
%	\item Lorem ipsum dolor sit amet
%	\item consectetur adipiscing elit.
%	\item Aliquam dignissim blandit est, in dictum tortor gravida eget. %In ac rutrum magna.
%\end{itemize}

%\section{Effect of Number of Linear %Combination within Unit and Depth}

In the numerical experiment, we compare our results of the maxout neural network Gaussian process (MNNGP) with the results of the finite-width, deep maxout network on the MNIST and CIFAR10 datasets. 
We also compare our MNNGPs with NNGPs in \cite{lee2017deep}. Then we have a discussion about the effect of hyperparameters in MNNGPs. For MNNGPs, we formulate the  classification problem as a regression task (\cite{rifkin2004defense}). That is, class labels are encoded as a one-hot, zero-mean, regression target (i.e., entries of -0.1 for the incorrect class
and 0.9 for the correct class). For training finite-width maxout networks, we will apply the mean square error (MSE) loss. The experiment setting of maxout networks with finite width can be found in \ref{finite_width}. And the setting of the Bayesian inference of MNNGPs can be founded in \ref{mnngp_inf}.

\subsection{Comparison with the Finite-width, Deep Maxout Network}

\begin{figure}[!tbp]
  \centering
  \subfloat[Comparison with the finite-width, maxout network on MNIST]{\includegraphics[width=0.9\textwidth, height=0.3\textwidth]{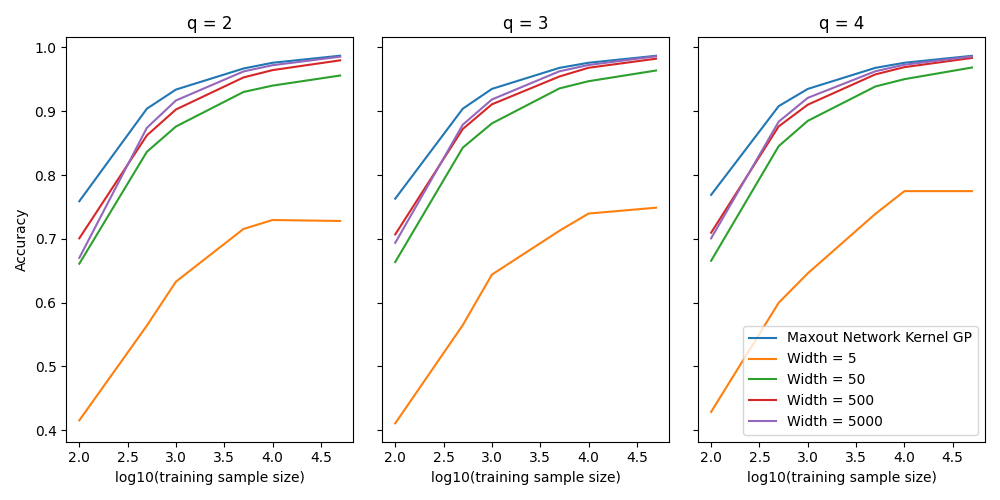}\label{fig1:f1}}

  \subfloat[Comparison with the finite-width, maxout network on CIFAR10]{\includegraphics[width=0.9\textwidth, height=0.3\textwidth]{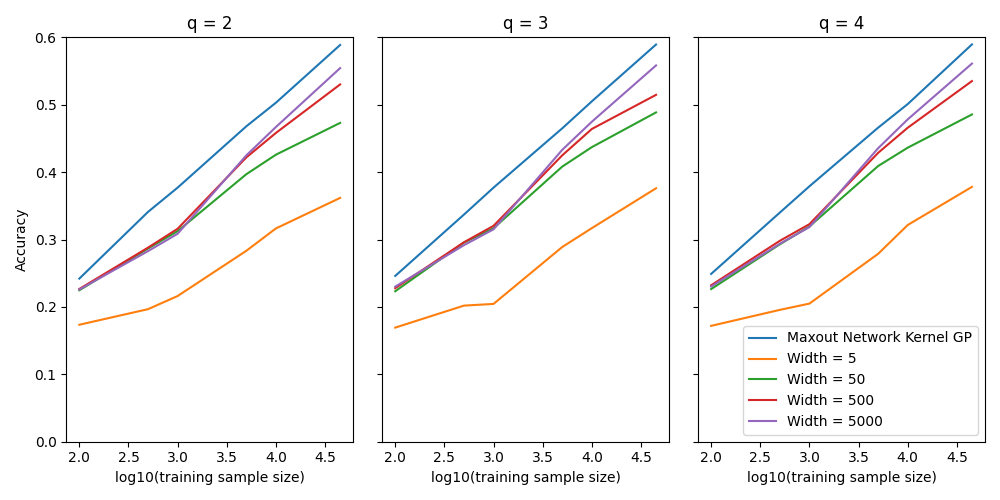}\label{fig2:f2}}
  \caption{For each combination (number of training sample, maxout rank, network width), the test accuracy corresponding to the best set of hyperparameters searched by grid search (see \ref{finite_width},\ref{mnngp_inf}) is reported.}
  \label{plots}
\end{figure}

Based on the experiment, no matter how large the maxout rank is, we can find that the Bayesian inference based on the MNNGP almost always outperforms their finite-width counterparts. Moreover, as the width of the maxout network gets larger, the performance of the finite-width, maxout network will be closer to that of the MNNGP. See Figure \ref{plots}.

\subsection{Comparison with NNGPs}
We find that our MNNGP have competitive results compared with NNGPs with ReLU and Tanh kernels on MNIST and CIFAR10 datasets. Especially, our Deep Maxout Network Kernel always outperforms NNGPs in the more challenging dataset CIFAR10 with a significant improvement when the size of training set is large enough. See Table ~\ref{table:1}. 

\begin{table}[!tbp]\centering
    \tiny
    \caption{\footnotesize The MNNGP often outperforms NNGPs with ReLU and Tanh avtivations from the view of test accuracies on MNIST and CIFAR-10 datasets. The reported MNNGP and NNGP results correspond to the best-performing combinations of tuning parameters on the validation set. Best models for a given training set size are specified by (activation-depth-$\sigma_{b}^{2}$-$\sigma_{w}^{2}$) for NNGPs, (widthq--maxout rank--depth-$\sigma_{b}$-$\sigma_{w}$) for finite-width maxout networks and (q-depth-$\sigma_{b}^{2}$-$\sigma_{w}^{2}$) for MNNGPs. $n_{t}$ is the size of the training set and $n_{v}$ is the size of the validation set.}
    \label{table:1}
    \begin{tabular}{ll}
    \begin{tabular}{|c|l|c|l|l|}
    \hline
    Dataset               & $n_{t}$              & \multicolumn{1}{l|}{$n_{v}$} &  Model             & Test accuracy \\ \hline
    \multirow{18}{*}{MNIST} & \multirow{4}{*}{100} & \multirow{18}{*}{10K}      & ReLU-100-0.83-1.79 & 0.7735        \\
                          &                      &                              & Tanh-100-0.97-3.14 & \textbf{0.7736}\\
                          &                      &                              & 500-3-13-0.1-0.001 & 0.7070         \\
                          &                      &                              & 4-9-1.17-3.48     & 0.7691        \\ \cline{2-2} \cline{4-5} 
                          & \multirow{4}{*}{500} &                              & ReLU-100-0.83-1.79 & 0.8995        \\
                         &                      &                              & Tanh-50-1.86-3.48  & 0.8277        \\
                          &                      &                              & 5000-4-9-0.1-0.001  & 0.8837       \\  
                          &                      &                              & 4-5-0.07-4.83     & \textbf{0.9079}  \\ \cline{2-2} \cline{4-5} 
                          & \multirow{4}{*}{1K}  &                              & ReLU-20-0.28-1.45  & 0.9279        \\
                          &                      &                              & Tanh-20-0.62-1.96  & 0.9266        \\
                          &                      &                              & 5000-4-21-0.1-0.001 & 0.9210        \\            
                          &                      &                              & 4-5-0.34-4.83     & \textbf{0.9350}  \\ \cline{2-2} \cline{4-5} 
                          & \multirow{4}{*}{5K}  &                              & ReLU-7-0.07-0.61   & 0.9692        \\
                          &                      &                              & Tanh-3-0.00-1.11   & \textbf{0.9693}\\
                          &                      &                              & 5000-4-1-0.1-0.001 & 0.9626        \\   
                          &                      &                              & 4-17-0.62-0.78      & 0.9683        \\ \cline{2-2} \cline{4-5} 
                          & \multirow{4}{*}{10K} &                              & ReLU-7-0.07-0.61   & 0.9765        \\
                          &                      &                              & Tanh-2-0.28-1.62   & \textbf{0.9773} \\
                          &                      &                              & 5000-4-5-0.1-0.001 & 0.9729       \\    
                          &                      &                              & 3-9-1.72-0.78     & 0.9763        \\ \cline{2-2} \cline{4-5} 
                          & \multirow{4}{*}{50K} &                              & ReLU-1-0.48-0.10   & 0.9875        \\
                          &                      &                              & Tanh-1-0.00-1.28   & \textbf{0.9879} \\
                          &                      &                              & 5000-4-1-0.5-0.001 & 0.9859       \\                        
                          &                      &                              & 2-5-1.72-1.45      & 0.9870        \\ \cline{2-2} \cline{4-5} 
    \hline
    \end{tabular}

    &

    \begin{tabular}{|c|l|c|l|l|}
    \hline
    Dataset               & $n_{t}$              & \multicolumn{1}{l|}{$n_{v}$} &  Model              & Test accuracy \\ \hline
  \multirow{18}{*}{CIFAR-10} & \multirow{4}{*}{100} &    \multirow{18}{*}{5K}& ReLU-3-0.97-4.49   & 0.2673 \\
    &                      &                              & Tanh-10-1.17-3.65  & \textbf{0.2718} \\
    &                      &                              & 500-3-21-0.1-0.001 & 0.2278         \\            
    &                      &                              & 4-13-0.07-2.13    &  0.2487          \\ \cline{2-2} \cline{4-5} 
    & \multirow{4}{*}{500} &                              & ReLU-20-0.21-1.79  & 0.3395        \\
    &                      &                              & Tanh-7-0.62-3.65   & 0.3291        \\
    &                      &                              & 500-3-21-0.1-0.001 & 0.2954         \\            
     &                      &                             &2-17-0.34-4.83     &\textbf{0.3410} \\ \cline{2-2} \cline{4-5} 
   & \multirow{4}{*}{1K}  &                              & ReLU-7-0.00-1.28   & 0.3608        \\
   &                      &                              & Tanh-50-0.97-2.97  & 0.3702        \\
    &                      &                              & 5000-4-9-0.1-0.001 & 0.3185        \\            
    &                      &                              & 4-17-0.07-4.16     & \textbf{0.3790} \\ \cline{2-2} \cline{4-5} 
     & \multirow{4}{*}{5K}  &                              & ReLU-3-1.03-4.66   & 0.4454        \\
     &                      &                              & Tanh-10-0.38-3.65  & 0.4430        \\
     &                      &                              & 5000-4-9-0.1-0.001 & 0.4355     \\            
   &                      &                              & 2-13-0.34-2.13       & \textbf{0.4677} \\ \cline{2-2} \cline{4-5} 
    & \multirow{4}{*}{10K} &                              & ReLU-5-0.28-2.97   & 0.4780        \\
     &                      &                              & Tanh-7-2.00-3.48   & 0.4766        \\
     &                      &                              & 5000-4-5-0.1-0.001 & 0.4783       \\
     &                      &                              & 3-21-0.07-1.45     &\textbf{0.5049}\\  \cline{2-2} \cline{4-5} 
    & \multirow{4}{*}{45K} &       & ReLU-3-1.86-3.31   & 0.5566 \\
     &                      & \multicolumn{1}{l|}{}        & tanh-3-1.52-3.48   & 0.5558        \\
    &                      &                              & 5000-4-1-0.1-0.001  & 0.5611  \\          
    &                      & \multicolumn{1}{l|}{}        & 4-13-0.34-4.16      & \textbf{0.5895}     \\        \hline
    \end{tabular}
    
    \end{tabular}
\end{table}

\normalsize

\subsection{Discussion about the Effect of Hyperparameters }

\begin{figure}[!tbp]
  \centering
  \subfloat[Maxout Rank q = 2]{\includegraphics[width=0.95\textwidth, height=0.37\textwidth]{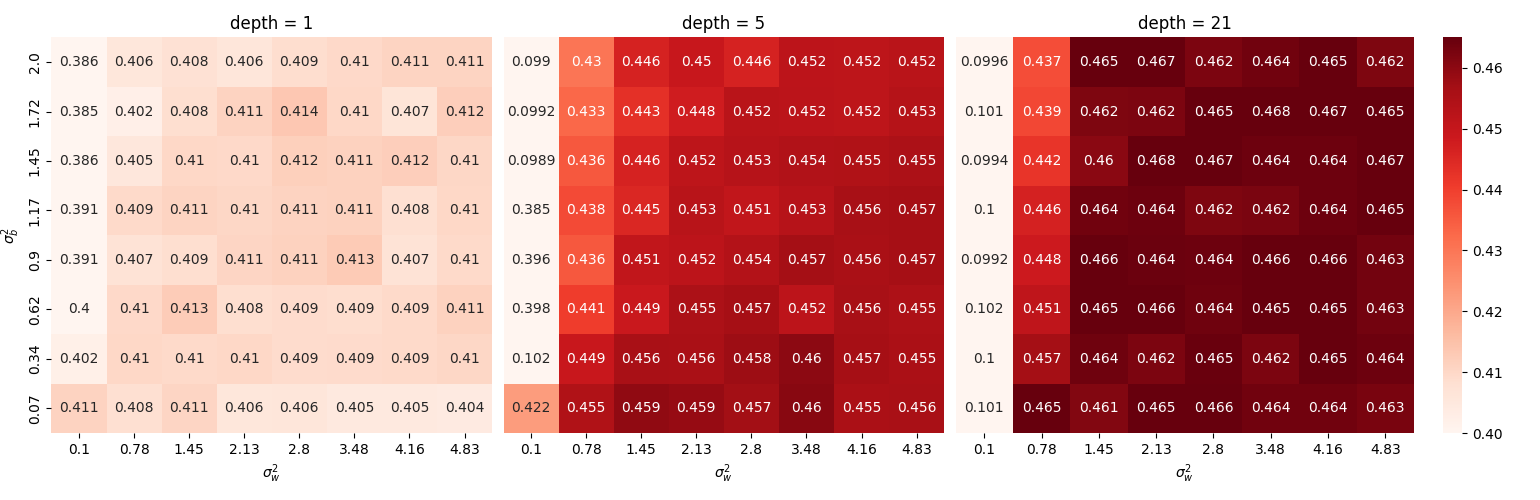}\label{fig3:f1}}

  \subfloat[Maxout Rank q = 3]{\includegraphics[width=0.95\textwidth, height=0.37\textwidth]{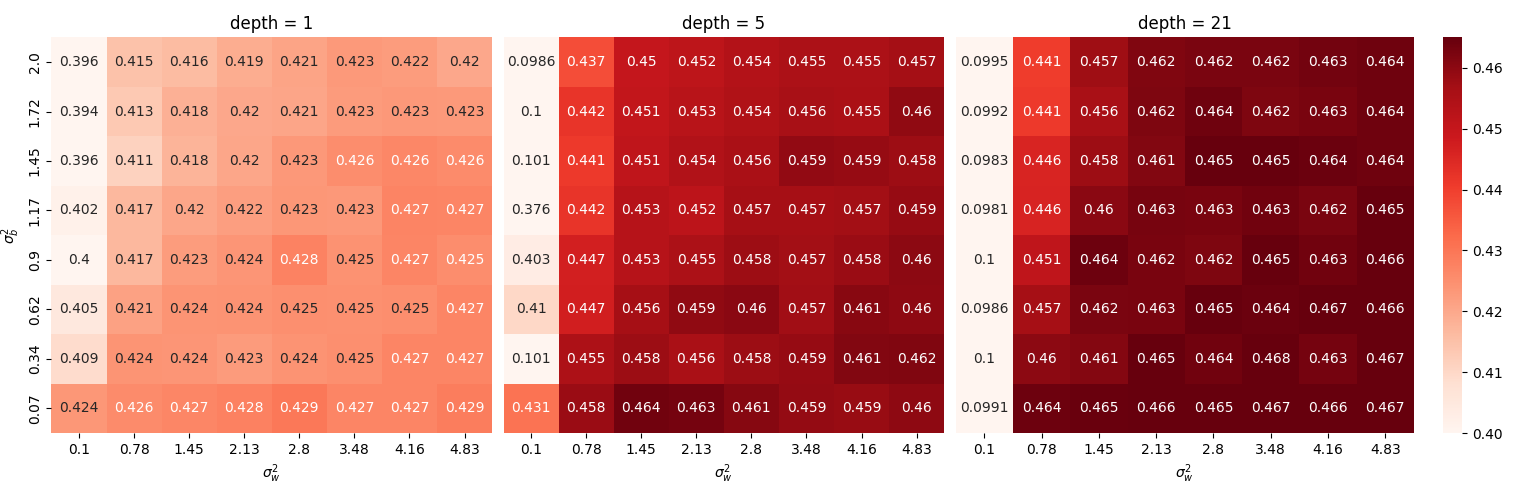}\label{fig3:f2}}
  
  \subfloat[Maxout Rank q = 4]{\includegraphics[width=0.95\textwidth, height=0.37\textwidth]{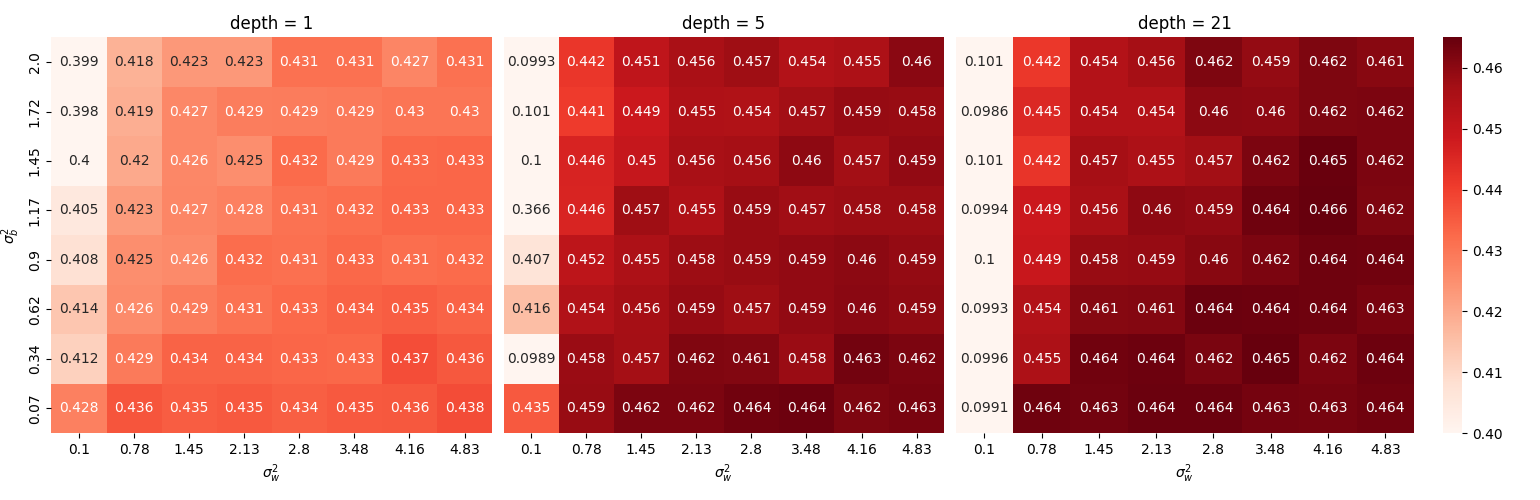}\label{fig3:f3}}
  
  \caption{The validation accuracy on CIFAR10 (see \ref{mnngp_inf}) for the training sample size = 5000}
  \label{plots2}
\end{figure}

For the MNNGP model, there are four main hyperparameters, the maxout rank $q$, the depth, $\sigma_w^2$ and $\sigma_b^2$. For the maxout rank $q$, we can find that larger maxout rank $q$ can lead to better performance when the network is shallow, while there is no big difference when the network is deep (see the 1st, 3rd column in Figure\ref{plots2}). For the depth, we can find that typically, larger depth can leads to better performance (see 1st, 2nd, 3rd row in Figure\ref{plots2}). For $\sigma_w^2$ and $\sigma_b^2$, when the the network gets deep, small $\sigma_w^2$ may lead to ill-conditioned $K_{X_{train},X_{train}}$, so that computing $K_{X_{train},X_{train}}^{-1}$ fails and generate poor predictions (see $\sigma_w^2=0.1$ in 3rd column of Figure\ref{plots2}). Generally, large $\sigma_w^2$ combined with moderate $\sigma_b^2$ can generate good performance.

\section{Discussion}

In this work, we derive the equivalence between the infinite-width, deep maxout network and the Gaussian process. The corresponding kernel is characterized by a compositional structure. One interesting discovery is that, with proper scale modification of the variance level of the initialization of biases and weights, as well as a proper scale of the input, the deep maxout network kernel with the maxout rank $q = 2$ is the same as the deep neural network kernel with the ReLU activation, which demonstrates the equivalence of the initial outputs of the two networks in the infinite-width regime. Whether the output of the two infinite-width networks will still be the same in the training process is another interesting topic for the future work. To apply the deep maxout network kernel, we propose an implementation based on numerical integration and interpolation which is efficient, with high precision and easy to be adapted to any maxout rank. Experiments on MNIST and CIFAR10 datasets show that Bayesian inference with the deep maxout network kernel can lead to competitive results compared with their finite-width counterparts and the results in \citet{lee2017deep}, especially in the more challenging dataset, CIFAR10. Currently, many infinite-width neural structures only consider nonlinear activations applied to single linear combination of the input such as the ReLU and Tanh (see \citet{novak2018bayesian}, \citet{garriga2018deep} and \citet{DBLP:journals/corr/abs-1910-12478}). We believe that incorporating the maxout activation into other infinite-width neural structures is promising to enhance the performances in practical application and this is left to future work. 

$\newline\newline$

\newpage
\bibliographystyle{unsrtnat}
\bibliography{references}  %%% Uncomment this line and comment out the ``thebibliography'' section below to use the external .bib file (using bibtex) .

%%% Uncomment this section and comment out the \bibliography{references} line above to use inline references.
% \begin{thebibliography}{1}

% 	\bibitem{kour2014real}
% 	George Kour and Raid Saabne.
% 	\newblock Real-time segmentation of on-line handwritten arabic script.
% 	\newblock In {\em Frontiers in Handwriting Recognition (ICFHR), 2014 14th
% 			International Conference on}, pages 417--422. IEEE, 2014.

% 	\bibitem{kour2014fast}
% 	George Kour and Raid Saabne.
% 	\newblock Fast classification of handwritten on-line arabic characters.
% 	\newblock In {\em Soft Computing and Pattern Recognition (SoCPaR), 2014 6th
% 			International Conference of}, pages 312--318. IEEE, 2014.

% 	\bibitem{hadash2018estimate}
% 	Guy Hadash, Einat Kermany, Boaz Carmeli, Ofer Lavi, George Kour, and Alon
% 	Jacovi.
% 	\newblock Estimate and replace: A novel approach to integrating deep neural
% 	networks with existing applications.
% 	\newblock {\em arXiv preprint arXiv:1804.09028}, 2018.

% \end{thebibliography}
\newpage
\appendix
%\section{Proof of Theorem 1}
\section{Technical Proofs}
\subsection{Proof of Theorem 1}
\label{proof}

\begin{proof}
The derivation will based on the Bayesian marginalization over intermediate layers described in \citet{lee2017deep}. 

Denote $X = \begin{pmatrix} \bold{x}^{\gamma_1}&\hdots&\bold{x}^{\gamma_B} \end{pmatrix}\in \mathbb{R}^{d_{in}\times B}$ and $Z_i=(z_i(\bold{x}^{\gamma_1}),...,z_i(\bold{x}^{\gamma_B}))^T\in\mathbb{R}^B$.
We will study the distribution of $Z_i|X$.

For each layer $l$, denote the second moment matrix as below:
\begin{equation}
\label{eq6}
K_{i,j}^l=\left\{
\begin{aligned}
&\frac{1}{d_{in}}\sum_{m=1}^{d_{in}}x^{\gamma_i}_m x^{\gamma_j}_m  , & l=0 \\
& \frac{1}{N_l}\sum_{m=1}^{N_l}x_m^l(\bold{x}^{\gamma_i}) x_m^l(\bold{x}^{\gamma_j}), & l>0  .
\end{aligned}
\right.\end{equation}

Note that since the weights and biases follow Gaussian distributions, we have
$Z_i|K^L \sim N(0, G(K^L))$.
From \eqref{15}, \eqref{16}, we know that $K^l$ only depend on $Z^{l-1}\equiv \{z_{m,b}^{l-1}(\bold{x}^{\gamma_{j}})\}$, where $Z^{l-1}$ denote the linear affine
combinations before the max-out activation in $(l-1)_{th}$ layer. And the distribution of $Z^{l-1}$ only depend on $K^{l-1}$ from the definition. So we have
\begin{equation}
    \begin{aligned}
        p(K^l|K^{l-1},K^{l-2},...,K^0,X)=P(K^l|K^{l-1}).
    \end{aligned}
\end{equation}

Define function $F,G$ as below:
\begin{equation}
    \begin{aligned}
        G(K^l)&=\sigma_w^2 K^l + \sigma_b^2\bold{1}\bold{1}^T \in \mathbb{R}^{B\times B}, \\
        H(K^l)&\in \mathbb{R}^{B\times B},
        ~where~~
        H(K^l)_{i,j}&= \sqrt{K^l_{i,i}}\sqrt{K^l_{j,j}}F_q(\frac{K^l_{i,j}}{\sqrt{K^l_{i,i}}\sqrt{K^l_{j,j}}}).
    \end{aligned}
\end{equation}
By the Law of Large Number (LLN), we could have for $l>0$,
\begin{equation}
    \begin{aligned}
        \lim_{N_l \to \infty}p(K^l|K^{l-1})=\delta(K^l-H \circ G(K^{l-1}) ),  ~~~~~~~\delta(s)=
%\label{eq6}
\left\{
\begin{aligned}
1 & , & s=0 \\
0& , & s\neq 0.
\end{aligned}
\right.
    \end{aligned}
\end{equation}

For $l=0$, obviously, we have
\begin{equation}
    \begin{aligned}
        p(K^0|X)=\delta(K^0-\frac{1}{d_{in}}X^TX).
    \end{aligned}
\end{equation}

Then we can write the $p(Z_i|X)$ as an integral over all the intermediate $K^l$, then we have
\begin{equation}
    \begin{aligned}
        \lim_{N_1 \to \infty,...,N_L\to\infty}p(Z_i|X)&=\lim_{N_1 \to \infty,...,N_L\to\infty}\int P(Z_i|K^L)*(\prod_{l=1}^Lp(K^l|K^{l-1}))*p(K^0|X)d K^0...dK^L\\
        &=\int p(Z_i|K^L)*(\prod_{l=1}^L\delta(K^l-H\circ G(K^{l-1})))*\delta(K^0-\frac{1}{d_{in}}X^TX)d K^0...dK^L \\
        &=p(Z_i|K^L= (H\circ G)^L(K^0)).
    \end{aligned}
\end{equation}

That is
\begin{equation}
    \begin{aligned}
      \lim_{N_1 \to \infty,...,N_L\to\infty}  Z_i|X \sim N(0, G\circ(H\circ G)^L(K^0)).
    \end{aligned}
\end{equation}
\end{proof}

\subsection{Proof of Proposition 1}
\label{pp}
\begin{proof}

We prove \textbf{Proposition 1} by induction. 

Let $F_{2}^{l}$ denote $F_2(\frac{\sigma_b^2+\sigma_w^2K^{l - 1}(\bold{x},\bold{x}^\prime)}{\sqrt{\sigma_b^2+\sigma_w^2K^{l - 1}(\bold{x},\bold{x})}*\sqrt{\sigma_b^2+\sigma_w^2K^{l - 1}(\bold{x}^\prime, \bold{x}^\prime)}})$ defined in \eqref{18}.

\textbf{STEP 1.}

For $l = 1$, eliminating subscript $k$ and superscript 0 in \eqref{5}, we have 
\begin{equation}
 F_{2}^{1} = \frac{1}{\sqrt{\text{Var} z_{1} (x) \cdot \text{Var} z_{1} (x^{\prime}) }} \mathbf{E}[\max(z_{1}(\bold{x}), z_{2}(\bold{x})) \cdot \max(z_{1}(\bold{x^{'}}), z_{2}(\bold{x^{'}}))]. \label{29}   
\end{equation}

Let $s, t, s^{'}, t^{'}$ denote $\frac{z_{1}(\bold{x})}{\sqrt{\text{Var} z_{1} (x)}}, \frac{z_{2}(\bold{x})}{\sqrt{\text{Var} z_{2} (x)}}, \frac{z_{1}(\bold{x^{'}})}{\sqrt{\text{Var} z_{1} (x^{\prime})}}, \frac{z_{2}(\bold{x^{'}})}{\sqrt{\text{Var} z_{2} (x^{\prime})}}$ in \eqref{29}, respectively. By \eqref{11}, we know  
\begin{align*}
\begin{pmatrix}s\\
t\\
s^\prime\\
t^\prime
\end{pmatrix} &\sim  N
\begin{bmatrix}
\begin{pmatrix}
0\\
0\\
0\\
0
\end{pmatrix}\;, &
\begin{pmatrix}
1 & 0 & \rho & 0 \\
0 & 1 & 0 & \rho\\
\rho & 0 & 1 & 0\\
0 & \rho & 0 & 1
\end{pmatrix}
\end{bmatrix}, \\[2\jot]
\end{align*}
where $\rho = \text{cor}{(z_{1}(x), z_{1} (x^{\prime}))}$. Then we have 
\begin{equation}
    \begin{aligned}
  F_{2}^{1} & = \mathbf{E}[\max(s, t) \cdot \max(s^{'}, t^{'})]  \\
     & = \mathbf{E}[(\max(s - t, 0) + t) \cdot (\max(s^{'} - t^{'}, 0) + t^{'})] \\
     & = \mathbf{E}[\max(s - t, 0)\cdot \max(s^{'} - t^{'}, 0)] + \mathbf{E}[\max(s - t, 0)\cdot t^{'}]  + \mathbf{E}[\max(s^{'} - t^{'}, 0)\cdot t] + \mathbf{E}[tt^{'}] \\
     & = \mathbf{E}[\max(s - t, 0)\cdot \max(s^{'} - t^{'}, 0)] + 2\mathbf{E}[\max(s - t, 0)\cdot t^{'}]  + \mathbf{E}[tt^{'}]\\
     & = \underbrace{\mathbf{E}[\max(\nu, 0)\cdot \max(\nu^{'}, 0)]}_{\cir{1}} + \underbrace{2 \mathbf{E}[\max(s - t, 0)\cdot t^{'}]  + \mathbf{E}[tt^{'}]}_{\cir{2}}\\
    \end{aligned}, \label{30}
\end{equation}
where 
\begin{align*}
\begin{pmatrix}
\nu\\
\nu^{\prime}
\end{pmatrix} &\sim  N
\begin{bmatrix}
\begin{pmatrix}
0\\
0\\
\end{pmatrix}\;, &
2
\begin{pmatrix}
1 & \rho \\
\rho & 1\\
\end{pmatrix}
\end{bmatrix}. \\[2\jot]
\end{align*}
Suppose we can show that $\cir{2} = 0$, then we have  
\begin{equation}
\begin{aligned}
    K^{1}(\bold{x}, \bold{x^{\prime}}) & =  \sigma_b^2+\sigma_w^2*\sqrt{\sigma_b^2+\sigma_w^2K^{0}(\bold{x},\bold{x})}*\sqrt{\sigma_b^2+\sigma_w^2K^{0}(\bold{x^{\prime}},\bold{x^{\prime}})}*\cir{1}  \\ 
    & = \Tilde{\sigma}_b^2 + \Tilde{\sigma}_w^2 * \sqrt{\frac{\text{Var}(z_{1}(x)) \cdot \text{Var}(z_{1}(x^{\prime}))}{4}} * \cir{1}\\
    & = \Tilde{\sigma}_b^2 + \Tilde{\sigma}_w^2 * \mathbf{E}[\max(z_{1}(x), 0)\cdot \max(z_{1}(x), 0)] \\
    & = K^{1}_{RL}(\frac{\bold{x}}{\sqrt{2}}, \frac{\bold{x^{\prime}}}{\sqrt{2}};\Tilde{\sigma}_{b}, \Tilde{\sigma}_{w}).
\end{aligned}
\end{equation}
Now we show that \cir{2} = 0. Since $s$, $t$ and $t^{'}$ are jointly normal, we have the following:
\begin{equation}
    \begin{aligned}
    \mathbf{E}[\max(s - t, 0) \cdot t^{\prime}] & = \mathbf{E}_{s - t}[\mathbf{E}[\max(s - t, 0) \cdot t^{\prime}| s - t]] \\
    & =  \mathbf{E}_{s - t}[\max(s - t, 0) \cdot \mathbf{E}[t^{\prime}| s - t]]\\
    & = -\frac{\rho}{2}\mathbf{E}_{s - t}[\max(s - t, 0) \cdot (s - t)]\\
    & = -\frac{\rho}{2}\mathbf{E}_{s - t}[(\max(s - t, 0))^{2}]\\
    & = -\frac{\rho}{2}. 
    \end{aligned}
\end{equation}
Thus, we have $\cir{2} = -\frac{\rho}{2} \cdot 2 + \rho = 0$.

%Actually, based on the \eqref{30} and $\cir{2}=0$, we can have $\forall \rho \in [-1,1]$
%\begin{equation}
%    F_2(\rho)=2*E[max\{0,k_1\}*max\{0,k_2\}]
%\end{equation}
%where $k_1\sim N(0,1), k_2\sim %N(0,1), cor(k_1,k_2)=\rho$.

\textbf{STEP 2.}

We suppose \eqref{19} holds for $l - 1$, $l > 1$, then for $l$
\begin{equation}
    \begin{aligned}
    K^l(\bold{x},\bold{x}^\prime) 
    & = \sigma_b^2+\sigma_w^2*\sqrt{\sigma_b^2+\sigma_w^2K^{l-1}(\bold{x},\bold{x})}*\sqrt{\sigma_b^2+\sigma_w^2K^{l-1}(\bold{x}^\prime,\bold{x}^\prime)}*F_q^{l} \\
    & = \sigma_b^2+\sigma_w^2*\sqrt{\sigma_b^2+\sigma_w^2K^{l-1}(\bold{x},\bold{x})}*\sqrt{\sigma_b^2+\sigma_w^2K^{l-1}(\bold{x}^\prime,\bold{x}^\prime)}*\mathbf{E}[\max(s, t) \cdot \max(s^{'}, t^{'})].  \\
    \end{aligned}
\end{equation}
Similarly as in \textbf{STEP 1.},
\begin{align*}
\begin{pmatrix}s\\
t\\
s^\prime\\
t^\prime
\end{pmatrix} &\sim  N
\begin{bmatrix}
\begin{pmatrix}
0\\
0\\
0\\
0
\end{pmatrix}\;, &
\begin{pmatrix}
1 & 0 & \rho^{'} & 0 \\
0 & 1 & 0 & \rho^{'}\\
\rho^{'} & 0 & 1 & 0\\
0 & \rho^{'} & 0 & 1
\end{pmatrix}
\end{bmatrix}, \\[2\jot]
\end{align*}
where $\rho^{'} = \frac{\sigma_b^2+\sigma_w^2K^{l-1}(\bold{x},\bold{x}^\prime)}{\sqrt{\sigma_b^2+\sigma_w^2K^{l-1}(\bold{x},\bold{x})}*\sqrt{\sigma_b^2+\sigma_w^2K^{l-1}(\bold{x}^\prime, \bold{x}^\prime)}}$. Noticing that the previous trick does not depend on the value of $\rho$, we have
\begin{equation}
\mathbf{E}[\max(s, t) \cdot \max(s^{'}, t^{'})] = \mathbf{E}[\max(\nu, 0)\cdot \max(\nu^{'}, 0)],   \label{34}
\end{equation}
where 
\begin{align*}
\begin{pmatrix}
\nu\\
\nu^{\prime}
\end{pmatrix} &\sim  N
\begin{bmatrix}
\begin{pmatrix}
0\\
0\\
\end{pmatrix}\;, &
2
\begin{pmatrix}
1 & \rho^{'} \\
\rho^{'} & 1\\
\end{pmatrix}
\end{bmatrix}. \\[2\jot]
\end{align*}
Then, by the induction assumption, we have 
\begin{equation}
    \begin{aligned}
     K^l(\bold{x},\bold{x}^\prime) & = \Tilde{\sigma}_b^2+ \Tilde{\sigma}_w^2*\sqrt{\sigma_b^2+\sigma_w^2K^{l-1}(\bold{x},\bold{x})}*\sqrt{\sigma_b^2+\sigma_w^2K^{l-1}(\bold{x}^\prime,\bold{x}^\prime)}*\mathbf{E}[\max(\nu, 0)\cdot \max(\nu^{'}, 0)]\\
     & = \Tilde{\sigma}_b^2+ \Tilde{\sigma}_w^2*\mathbf{E}_{z^{l - 1} \sim \mathcal{GP}(0, K^{l - 1}(x, x^{\prime}))}[\max(z^{l - 1}(x), 0)\cdot \max(\max(z^{l - 1}(x^{'}), 0)] \\
     & = \Tilde{\sigma}_b^2+ \Tilde{\sigma}_w^2*\mathbf{E}_{z^{l - 1} \sim \mathcal{GP}(0, K^{l - 1}(\frac{\bold{x}}{\sqrt{2}}, \frac{\bold{x}^{\prime}}{\sqrt{2}};\Tilde{\sigma}_b^2, \Tilde{\sigma}_w^2))}[\max(z^{l - 1}(x), 0)\cdot \max(\max(z^{l - 1}(x^{'}), 0)\\
     & =  K^{l}(\frac{x}{\sqrt{2}}, \frac{x^{\prime}}{\sqrt{2}};\Tilde{\sigma}_b^2, \Tilde{\sigma}_w^2)
     . \\
    \end{aligned}
\end{equation}
Thus, \eqref{19} also holds for $l$ when it holds for $l - 1$ . We finish the proof. 
\end{proof}

\section{Evaluation of $F_q(\rho)$}
%(mention parameter $n_{\rho}$)
$\newline$
Let $\phi(x)$ and $\Phi(x)$ denote the pdf and cdf of $N(0,1)$, respectively; $\phi_{2,\rho}(x,y)$ and $\Phi_{2,\rho}(x,y)$ denote the pdf and cdf of $N_2\bigg(0, \begin{pmatrix}1,\rho\\ \rho,1 \end{pmatrix}\bigg)$. Let $R_{max}$ denote the maximum integration range and $n_{grid}$ the number of grid, then we take $n_{grid}\times n_{grid}$ points from the range $[-R_{max}, R_{max}]\times [-R_{max}, R_{max}]$, which are denoted by $\{(x_i, y_j)\}_{i,j=1,...,n_{grid}}$. $\phi(x)$,  $\Phi(x)$,  $\phi_{2,\rho}(x,y)$, and $\Phi_{2,\rho}(x,y)$ can be evaluated with function $scipy.stats.multivariate\_normal$\href{https://docs.scipy.org/doc/scipy/reference/generated/scipy.stats.multivariate_normal.html}{[link]} and $scipy.stats.norm$\href{https://docs.scipy.org/doc/scipy/reference/generated/scipy.stats.norm.html}{[link]} in Python easily.

$\newline$
Here we first introduce how to evaluate $F_q(\rho)$ for the grids $\{\rho_i\}_{i=1,...,n_\rho}\subset [-1,1]$. \label{implementation}

\subsection{Evaluate $F_q(\rho)$ for  $\rho\in(-1,1)$}

Based on the definition in \eqref{4}, we denote $I=\arg \max_{i}\{h_i\}$ and $I^\prime = \arg \max_{i}\{h_i^\prime\}$. Then, we have 
\begin{equation}
    \begin{aligned}
      E[max\{h_1,...,h_q\}*max\{h_1^\prime,...,h_q^\prime\}]&=E[max\{h_1,...,h_q\}*max\{h_1^\prime,...,h_q^\prime\}*\bold{1}_{I=I^\prime}]\\
      &~~~~+E[max\{h_1,...,h_q\}*max\{h_1^\prime,...,h_q^\prime\}*\bold{1}_{I\neq I^\prime}] \\
      &=q*E[max\{h_1,...,h_q\}*max\{h_1^\prime,...,h_q^\prime\}*\bold{1}_{I=I^\prime=1}] \\
      &~~~~+q*(q-1)E[max\{h_1,...,h_q\}*max\{h_1^\prime,...,h_q^\prime\}*\bold{1}_{I=1,I^\prime=2}].
    \end{aligned}
\end{equation}

Thus, we can focus on the evaluation of $E[max\{h_1,...,h_q\}*max\{h_1^\prime,...,h_q^\prime\}*\bold{1}_{I=I^\prime=1}]$ and $E[max\{h_1,...,h_q\}*max\{h_1^\prime,...,h_q^\prime\}*\bold{1}_{I=1,I^\prime=2}]$

\subsubsection{Evaluate $E[max\{h_1,...,h_q\}*max\{h_1^\prime,...,h_q^\prime\}*\bold{1}_{I=I^\prime=1}]$}

From the definition in \eqref{4}, we know that the set of pairs $\{(h_i,h_i^\prime)\}_{i=1,...,q}$ are i.i.d. random pairs with distribution $N_2\bigg(0, \begin{pmatrix}1,\rho\\ \rho,1 \end{pmatrix}\bigg)$. Then we have
\begin{equation}
    \begin{aligned}
      E[max\{h_1,...,h_q\}*max\{h_1^\prime,...,h_q^\prime\}*\bold{1}_{I=I^\prime=1}]&=\int xy\phi_{2,\rho}(x,y)\Phi^{q-1}_{2,\rho}(x,y) dx dy.
    \end{aligned}
\end{equation}
Applying numerical integration. we have
\begin{equation}
    \begin{aligned}
      \int xy\phi_{2,\rho}(x,y)\Phi^{q-1}_{2,\rho}(x,y) dx dy&\approx \frac{\sum_{i,j=1}^{n_{grid}}x_iy_j\phi_{2,\rho}(x_i,y_j)\Phi^{q-1}_{2,\rho}(x_i,y_j)}{\sum_{i,j=1}^{n_{grid}}\phi_{2,\rho}(x_i,y_j)\Phi^{q-1}_{2,\rho}(x_i,y_j)}.
    \end{aligned}
\end{equation}

\subsubsection{Evaluate $E[max\{h_1,...,h_q\}*max\{h_1^\prime,...,h_q^\prime\}*\bold{1}_{I=1,I^\prime=2}]$}

From the definition in \eqref{4}, we have 
\begin{equation}
    \begin{aligned}
        (h_1,h_1^\prime, h_2, h_2^\prime)&\sim N_4\bigg(0,\begin{pmatrix}1&\rho&0&0\\\rho&1&0&0\\ 0&0&1&\rho\\0&0&\rho&1\end{pmatrix}\bigg).
    \end{aligned}
\end{equation}
Then, we have
\begin{equation}
    \begin{aligned}
       (h_1^\prime, h_2)|(h_1=x, h_2^\prime = y)\sim N_2\bigg( \begin{pmatrix}\rho*x\\\rho*y \end{pmatrix}, \begin{pmatrix}1-\rho^2&0\\0&1-\rho^2 \end{pmatrix}\bigg).
    \end{aligned}
\end{equation}
So,
\begin{equation}
    \begin{aligned}
       E[max\{h_1,...,h_q\}*max\{h_1^\prime,...,h_q^\prime\}*\bold{1}_{I=1,I^\prime=2}]&=\int xy\phi(x)\phi(y)\Phi(\frac{(1-\rho)x}{\sqrt{1-\rho^2}})\Phi(\frac{(1-\rho)y}{\sqrt{1-\rho^2}})\Phi_{2,\rho}(x,y)dx dy .\\
    \end{aligned}
\end{equation}
Applying numerical integration, we have
\small
\begin{equation}
    \begin{aligned}
        \int xy\phi(x)\phi(y)\Phi(\frac{(1-\rho)x}{\sqrt{1-\rho^2}})\Phi(\frac{(1-\rho)y}{\sqrt{1-\rho^2}})\Phi^{q-2}_{2,\rho}(x,y)dx dy &\approx \frac{\sum_{i,j=1}^{n_{grid}}  x_iy_j\phi(x_i)\phi(y_j)\Phi(\frac{(1-\rho)x_i}{\sqrt{1-\rho^2}})\Phi(\frac{(1-\rho)y_j}{\sqrt{1-\rho^2}})\Phi_{2,\rho}^{q-2}(x_i,y_j)}{\sum_{i,j=1}^{n_{grid}} \phi(x_i)\phi(y_j)\Phi(\frac{(1-\rho)x_i}{\sqrt{1-\rho^2}})\Phi(\frac{(1-\rho)y_j}{\sqrt{1-\rho^2}})\Phi_{2,\rho}^{q-2}(x_i,y_j)}.
    \end{aligned}
\end{equation}
\normalsize

\subsection{Evaluate $F_q(\rho)$ for $\rho = \pm 1$}

For $\rho = 1$, $h_i=h_i^\prime$ almost surely. Then, we have
\begin{equation}
    \begin{aligned}
      E[max\{h_1,...,h_q\}*max\{h_1^\prime,...,h_q^\prime\}]&=E[(max\{h_1,...,h_q\})^2] \\
      &=q*\int x^2 \phi(x)\Phi^{q-1}(x)dx\\
      &\approx q*\frac{\sum_{i=1}^{n_{grid}} x_i^2 \phi(x_i)\Phi^{q-1}(x_i)}{\sum_{i=1}^{n_{grid}} \phi(x_i)\Phi^{q-1}(x_i)}. \\
    \end{aligned}
\end{equation}

For $\rho=-1$, $h_i=-h_i^\prime$ almost surely. Then, we have
\small
\begin{equation}
    \begin{aligned}
      E[max\{h_1,...,h_q\}*max\{h_1^\prime,...,h_q^\prime\}]&=-E[(max\{h_1,...,h_q\})*min\{h_1,...,h_q\}] \\
      &=-q(q-1)*\int_{x>y} xy\phi(x)\phi(y)(\Phi(x)-\Phi(y))^{q-2}dx dy\\
      &\approx -q(q-1)\frac{\sum_{i=1}^{n_{grid}} x_iy_j\phi(x_i)\phi(y_j)*(\Phi(x_i)-\Phi(y_j))^{q-2}*\bold{1}_{x_i>y_j} }{\sum_{i=1}^{n_{grid}} \phi(x_i)\phi(y_j)(\Phi(x_i)-\Phi(y_j))^{q-2}*\bold{1}_{x_i>y_j}}.\\
    \end{aligned}
\end{equation}
\normalsize

Then, for $\forall \rho \in [-1,1]$, $F_q(\rho)$ will be evaluated by interpolation based on the grids.

\subsection{Evaluation of $F_q(\rho)$ by Interpolation}

For $\forall \rho \in [-1,1]$, let $\rho \in [\rho_i,\rho_{i+1}]$, then, we have
\begin{align*}
    F_q(\rho) = \frac{\rho_{i+1}-\rho}{\rho_{i+1}-\rho_i} *F_q(\rho_i) + \frac{\rho-\rho_{i}}{\rho_{i+1}-\rho_i} *F_q(\rho_{i+1}).
\end{align*}

\section{Details of Experiment Setup}

\subsection{Setup for the finite width, deep maxout network}

\textbf{Hyperparameters tuning}

The learning rate of training finite-width, deep maxout networks is $10^{-5}$. The batch size is 256. The number of epochs is 200.

We apply grid search to tune other hyperparameters. We select the width from $\{ 5, 50, 500, 5000 \}$, maxout rank $q$ from $\{ 2, 3, 4 \}$, depth from $\{ 1, 5, 9, 13, 17, 21 \}$, the standard deviation of the initialization of weights from $\{0.001, 0.01, 0.1, 0.5, 1\}$, the standard deviation of the initialization of biases from $\{0.1, 0.5, 1\}$.

For each choice of hyperparameters and for both MNIST and CIFAR-10 datasets, the validation results and testing results are generated the same as \ref{mnngp_inf}.

\label{finite_width}

\subsection{Setup for the Bayesian inference based on MNNGP}

\label{mnngp_inf}

\textbf{Parameters of Lookup Tables}

For all the experiments we use pre-computed lookup tables
F with $n_{grid}$ = 501, $n_{\rho}$ = 1001, and $R_{max}$ = 100. The target noise $\sigma_\epsilon^{2}$ is set to $10^{-10}$ initially, and is increased by a factor of 10 when the Cholesky decomposition failed while
solving \eqref{20}.

\textbf{Hyperparameters tuning}

We select $q$ from $\{ 2, 3, 4 \}$, depth from $\{ 1, 5, 9, 13, 17, 21 \}$, $\sigma_{b}^{2}$ from $\{ \frac{2 \cdot i}{29} |\; i \text{ mod } 4 = 1, i \in [30]\}$ and $\sigma_{w}^{2}$ from $\{ 0.1 + \frac{49 \cdot i}{290} | \; i \text{ mod } 4 = 0, i \in [30]\}$.

\textbf{Validation Results Report}

For both MNIST and CIFAR-10 datasets, we randomly shuffle the original training set first. Then the first $n_{t}$ samples from the shuffled training set are selected to be the training set and first $n_{v}$ samples from the remaining ones are treated as the validation sets. For each combination of hyperparameters, we estimate the model on the training set and calculate validation accuracy on the validation set. We repeat the previous procedure 20 times and choose the combination of hyperparameters with the highest mean accuracy.         

\textbf{Testing Results Report}
After selecting hyperparameters, we apply the previous data splitting procedure on the training set, estimate the model on the training set of size $n_{t}$ and calculate the testing accuracy on the original testing set. We repeat the procedure 20 times and report the mean testing accuracy.   

\end{document}